\documentclass[letterpaper, 10 pt, conference]{ieeeconf}  % Comment this line out if you need a4paper

\IEEEoverridecommandlockouts  % This command is only needed if you want to use the \thanks command

\usepackage{booktabs}       % professional-quality tables
\usepackage{amsfonts}       % blackboard math symbols
\usepackage{nicefrac}       % compact symbols for 1/2, etc.
\usepackage{microtype}      % microtypography
\usepackage{xcolor}         % colors
\usepackage{graphicx}
\usepackage{amsmath}
\usepackage{amssymb}
\usepackage{algorithm}
\usepackage{algorithmic}
\usepackage{tabularx}
\usepackage{longtable}
\usepackage[table,xcdraw]{xcolor}
\usepackage{subcaption}

\title{\LARGE \bf
Pseudocode-Guided Structured Reasoning for Automating Reliable Inference in Vision-Language Models
}

\author{
Weicong Ni, Tianbao Jiang and Linlin Wang$^*$
\thanks{All authors are with East China Normal University, Shanghai, China.
        {\tt\small [51265901038, tbjiang]@stu.ecnu.edu.cn}}%
\thanks{$^{*}$Corresponding author: Linlin Wang (E-mail: llwang@cs.ecnu.edu.cn)}%
\thanks{This work was supported by New Generation Artificial Intelligence-National Science and Technology Major Project (Grant No. 2025ZD0123402), the Computational Biology Program (Grant No. 25JS2830402) of Science and Technology Commission of Shanghai Municipality (STCSM), and Shanghai Municipal Science and Technology Major Project (Grant No. 2025SHZDZX025G06). We also thank the financial support from Shanghai Chinafortune Co., Ltd.}%
}

\begin{document}

\maketitle
\thispagestyle{empty}
\pagestyle{empty}

%%%%%%%%%%%%%%%%%%%%%%%%%%%%%%%%%%%%%%%%%%%%%%%%%%%%%%%%%%%%%%%%%%%%%%%%%%%%%%%%
\begin{abstract}

Vision-Language Models (VLMs) are becoming the cornerstone of high-level reasoning for robotic automation, enabling robots to parse natural language commands and perceive their environments. However, their susceptibility to hallucinations introduces critical failures in decision-making, posing significant safety and reliability risks in physical deployments. This challenge is exacerbated by the open-ended nature of real-world tasks, where questions vary vastly in difficulty and modality, demanding robust and adaptable reasoning strategies.
To tackle this, we propose the \underline{P}seudocode-guided \underline{St}ructured Re\underline{a}soning f\underline{r}amework (PStar), which adaptively selects structured pseudocode reasoning paths to help VLMs perform flexible and step-by-step reasoning.
We first design a set of abstract reasoning functions and formulate a structured pseudocode library to represent modular reasoning strategies.
Crucially, we design a Difficulty Feature Vector (DFV) that allows the model to assess question complexity and adaptively choose appropriate reasoning strategies—enhancing robustness and interpretability.
Extensive experiments demonstrate that PStar significantly reduces hallucination rates, achieving state-of-the-art scores of 87.1\% on POPE and 68.0\% on MMStar, outperforming even GPT-4V.
By providing a validated mechanism to reduce visual-language errors, PStar offers a critical step toward deploying more trustworthy and deterministic VLMs for real-world automated systems, where such errors can lead to catastrophic outcomes.

\end{abstract}

%%%%%%%%%%%%%%%%%%%%%%%%%%%%%%%%%%%%%%%%%%%%%%%%%%%%%%%%%%%%%%%%%%%%%%%%%%%%%%%%
\section{Introduction}
\label{sec:intro}

Vision-Language Models (VLMs) have made remarkable progress in robotic and automatic tasks such as visual question answering (VQA) in robotic surgery \cite{robot_vqa_surgery} and human-object interaction \cite{automatic_HOI}.
However, their deployment in safety-critical automation domains is severely limited by a fundamental and unacceptable risk, namely hallucination, where the model generates incorrect or fabricated information that leads to catastrophic failures \cite{MiniGPT-4}.
Eliminating such hallucinations is therefore a critical prerequisite for achieving the reliable and trustworthy reasoning required in safety-critical systems.

Existing approaches to mitigate hallucinations remain insufficient for robotic automation. Conventional methods such as tool augmentation \cite{wang2024cogvlm} and training alignment \cite{yu2024rlaifv} lack the adaptability needed in open-world settings.
Reasoning-based techniques offer more structured thinking but introduce new bottlenecks.
For instance, sequential Chains-of-Thought \cite{LLAVA-COT, DeepSeek} are computationally efficient but prone to a critical flaw: error propagation. This phenomenon, known as the hallucination snowball effect, occurs when an initial mistake in perception (e.g., failing to detect an obstacle) invalidates all subsequent reasoning and planning steps, guaranteeing that the robot's final action will be incorrect and potentially unsafe \cite{Hallucination_Snowball}.
Tree-based methods \cite{AR-MCTS, AStar} explore multiple reasoning paths but are too slow for real-time robotic response.
Code-guided frameworks \cite{SBSC} improve grounding but are inflexible across perceptual tasks requiring nuanced vision-language integration, such as understanding spatial and complex relationships.

What is urgently needed is a reasoning framework that is not only accurate and reliable but also efficient and adaptable-one that dynamically aligns reasoning effort with the complexity of the perceptual question, thereby ensuring real-time viability and safety in automated systems. 
To tackle these challenges, we introduce \textbf{P}seudocode-guided \textbf{St}ructured Re\textbf{a}soning f\textbf{r}amework (\textbf{PStar}), a novel framework that enhances multimodal reasoning by leveraging structured pseudocode reasoning paths.
Given that questions of varying difficulty require different reasoning paths, we propose the Difficulty Feature Vector (DFV) to adaptively select the most appropriate reasoning path. 
PStar constructs a pseudocode library containing diverse reasoning paths, which facilitates structured reasoning to effectively reduce hallucinations.

Our main contributions are as follows:
\begin{itemize}
    \item We propose \textbf{PStar}, a novel reasoning framework that uses structured pseudocode to make Vision-Language Model (VLM) reasoning interpretable, reliable, and adaptable, which are critical properties for deploying VLMs in safety-sensitive robotic tasks such as action planning and human instruction following.
    \item We introduce a modular pseudocode library of abstract reasoning functions and a Difficulty Feature Vector (DFV) that together allow robots to dynamically adjust their reasoning strategy based on perceived task complexity. This provides a principled approach to minimize overthinking under simple conditions and activate deeper reasoning in complex scenes.
    \item We demonstrate that PStar significantly reduces hallucination and improves reasoning robustness in visually complex and linguistically varied settings. Our method enables qwen2.5-VL-7B to outperform GPT-4V on curated benchmarks (68.0 on MMStar, 87.1 on POPE), offering a lightweight yet powerful solution for reliable automation.
\end{itemize}

\begin{figure*}[t]
    \centering
    \includegraphics[width=1.0\linewidth]{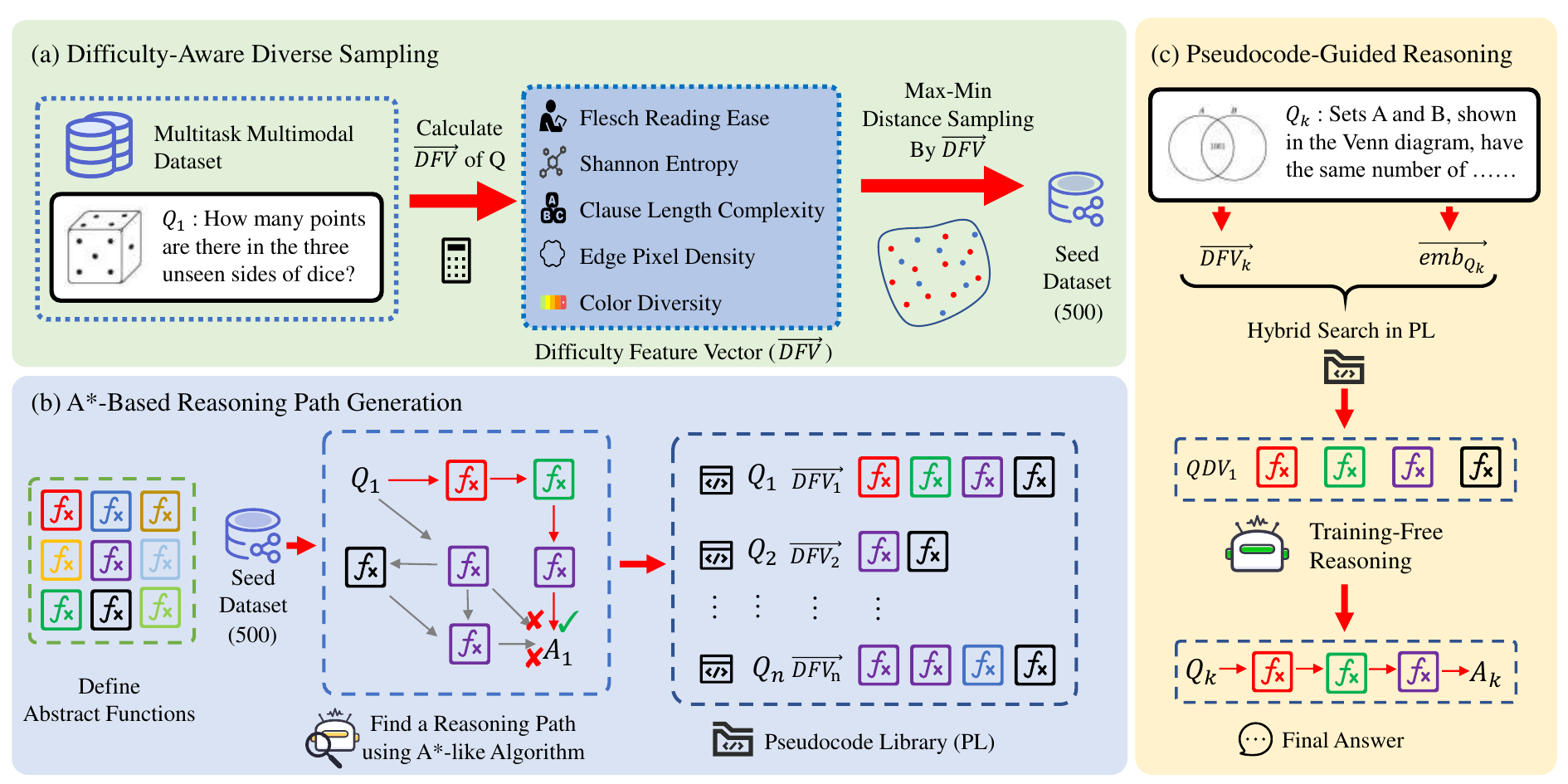}
    \caption{Overview of PStar for Structured Reasoning. (a) Difficulty-Aware Sampling uses Difficulty Feature Vectors (DFV) and max-min sampling to ensure diversity and relevance; (b) A*-Based Path Generation leverages A* search with an LVLM to discover reasoning paths; (c) Pseudocode-guided Reasoning applies selected paths to guide structured reasoning.}
    \label{fig:main}
\end{figure*}

\section{Methodology}
\label{sec:method}
\subsection{Motivation and Model Overview}
Multimodal reasoning in vision-language tasks suffers from a critical misalignment: existing methods lack adaptive mechanisms to match reasoning strategies with varying question complexity. This leads to either oversimplified reasoning that fails to capture nuanced details, or excessively complex paths that introduce unnecessary computational overhead and instability \cite{Overthinking_Survey, Underthinking}.
In robotic automation contexts where visual reasoning directly informs physical actions, this mismatch becomes particularly critical, as unstructured reasoning processes frequently result in hallucinations that compromise system reliability and safety.

To address these challenges, we propose the \textbf{P}seudocode-guided \textbf{St}ructured Re\textbf{a}soning f\textbf{r}amework (\textbf{PStar}). Our approach incorporates three key components (Figure~\ref{fig:main}):
(a) Difficulty-Aware Diverse Sampling, which utilizes Difficulty Feature Vectors (DFV) and Max-Min Distance Sampling to create a diverse seed dataset covering various difficulty levels;
(b) A*-Based Reasoning Path Generation, which combines A* search with an LVLM to generate question-specific reasoning paths;
(c) Pseudocode-guided Reasoning, which directs the model through these structured paths without additional training, effectively reducing hallucinations.
% Together, these components form a unified framework that adapts to question complexity and mitigates hallucinations through training-free reasoning.

\subsection{Difficulty-Aware Diverse Sampling}
\label{sec:dataset}
\paragraph{Dataset Construction}
To comprehensively cover and sample questions with diverse difficulty features, we construct a mixed multimodal reasoning dataset by integrating three datasets: \textbf{MathVista} \cite{MathVista}, \textbf{MATH-Vision} \cite{MathVision}, and \textbf{ScienceQA} \cite{ScienceQA}.
MathVista provides mathematical problems with visual contexts, MATH-Vision covers 16 mathematical disciplines, and ScienceQA includes multimodal science questions requiring multi-hop reasoning. This combination results in a diverse dataset of 10,258 questions across multiple formats, ensuring broad coverage of reasoning types and modalities.

\paragraph{Difficulty Feature Vector}
To account for varying question difficulties, we introduce the Difficulty Feature Vector (DFV), which quantifies complexity across textual and visual dimensions.
DFV includes five metrics: Flesch Reading Ease, Shannon Entropy, Clause Length Complexity, Edge Pixel Density and Color Diversity.
Flesch Reading Ease (FRE) \cite{FRE} measures text accessibility through the formula $206.835 - 1.015 \times \text{ASL} - 84.6 \times \text{ASW}$, where ASL (average sentence length) and ASW (average syllables per word) quantify linguistic complexity.
We use Clause Length Complexity (CLC = ASL) to represent syntactic structure, since ASL effectively captures clause-level complexity patterns.
Shannon Entropy evaluates lexical diversity, is computed as $-\sum_x p(x) \log p(x)$, where p(x) is the frequency (or empirical probability) of word $x$ within the given text.
For the visual modality, Edge Pixel Density is defined as the ratio of the number of edge pixels to the total number of pixels in the image. It is computed using the Canny edge detector \cite{Canny}.
Color Diversity is computed as the ratio of the number of pixels with unique RGB values to the total number of pixels in the image.
We apply z-score normalization to standardize these features.
This multidimensional approach provides a comprehensive and interpretable framework for assessing multimodal question difficulty.

\paragraph{Max-Min Distance Sampling}
To support diverse reasoning paths with A*-based search, we select a seed dataset that spans a wide range of multimodal question complexities quantified by the DFV.
We apply the Max-Min distance sampling algorithm to select 500 seed questions that are maximally diverse in the DFV space. 
This strategically ensures broad coverage of reasoning patterns while maintaining computational efficiency, preserving the method's generalization capability without restricting solution diversity.

\subsection{A*-Based Reasoning Path Generation}
\label{sec:af}
The limitations of long chain-of-thought (CoT) reasoning, including tendencies toward overthinking or underthinking \cite{Overthinking_Survey, Underthinking}, motivate our integration of adaptive thinking strategies.

\paragraph{Abstract Functions Definition}
Inspired by code planning \cite{CODEIO} and natural language planning \cite{AStar}, we introduce a pseudocode planning framework that employs abstract functions to guide the reasoning process of vision-language models (VLMs). These functions are categorized by modality into textual and visual operations.
We define abstract functions that build upon insights from \cite{AStar}, including \texttt{Visual Analysis (VA)}, \texttt{System Analysis (SA)}, \texttt{Regular Reasoning (RR)}, \texttt{Self-Reflection (SR)}, and \texttt{Numerical Analysis (NA)}.
Additionally, inspired by \cite{Atom_of_Thoughts} and \cite{Overthinking_Survey}, we introduce Simplify Problem (SP), Knowledge Injection (KI), Output Answer (OA), and Error Reasoning (EA), respectively.
By defining a set of abstract decision functions, pseudocode planning provides a mechanism to explicitly control and adjust the reasoning strategies of VLMs. This approach combines the formal precision of programming languages with the flexibility of natural language, enabling more structured, adaptable, and reliable reasoning.
% A detailed description of these abstract functions is provided in the Appendix \ref{sec:def_of_af}.

\paragraph{A* Search for Reasoning Paths}
\label{sec:astar}
We introduce an A*-based framework to reduce hallucinations in multimodal models by prioritizing reasoning steps according to abstract reasoning functions. This approach offers determinism, in contrast to random methods like MCTS \cite{AR-MCTS}. The A* algorithm generates a reasoning path for each question, composed of abstract reasoning functions, and stores them in a pseudocode library to create a structured framework for multimodal reasoning.
Inspired by Levin Tree Search \cite{Levin_Tree_Search} and its formula, we propose the following novel cost function:
\begin{equation}
\label{eq:g_n_ours}
g(S) =\sum_i \frac{\lambda_{a_i} \cdot \text{len}(r_i)}{\text{usefulness}(r_i)}
\end{equation}
\noindent
where \(a_i\) denote the \(i\)-th abstract reasoning function and \(r_i\) its corresponding response.
Each function \(a\) is assigned a cost coefficient \(\lambda_a\), with \(\lambda_{RR} = 1\) for regular reasoning steps directly contributing to the answer, and \(\lambda_a > 1\) for auxiliary steps such as analysis or reflection.
The usefulness function \(\text{usefulness}(r_i)\) measures the proportion of novel tokens in \(r_i\) relative to all prior responses \(\{r_j \mid j < i\}\), encouraging information diversity to enrich context and improve answer accuracy.
% A full list of cost coefficients is provided in Appendix~\ref{sec:coefficient}.

The heuristic cost function $h(S, a)$ is defined as follows:

\begin{equation}
h(S, a) = \lambda_a \cdot \text{len}(r) + \frac{\lambda_{RR} \cdot  (M - G - \text{len}(r))}{\text{usefulness}(r')}
\end{equation}

\noindent where \(M\) is the expected total number of tokens, and \(G\) is the number of tokens already generated.  \(a\) represents the next abstract reasoning function, with \(\text{len}(r)\) as the maximum token limit for response \(r\).
\(r'\) is the sequence of regular reasoning \texttt{RR()} responses  after applying abstract function \(a\), continuing until the reasoning reaches the goal node.

In the A* algorithm, the heuristic cost function \(h(S, a)\) estimates the cost of reaching the goal by applying abstract reasoning function \(a\) from state \(S\).
After applying \(a\), the model proceeds with regular reasoning (\(RR\)) steps to approximate the shortest path.
For simplicity, we assume these \(RR\) steps are maximally useful (\(\text{usefulness}(r') = 1\)) with cost \(\lambda_{RR} = 1\).
The total cost is thus the sum of the immediate and estimated future costs:

\begin{align}
\label{eq:f_s_all}
\nonumber
f(S,a) = g(S) + h(S,a) = \sum_i \frac{\lambda_{a_i} \cdot \text{len}(r_i)}{\text{usefulness}(r_i)} + \\ \lambda_a \cdot \text{len}(r) + (M - G - \text{len}(r))
\end{align}

Our search process terminates immediately upon generating a correct answer. To ensure efficiency, we enforce strict constraints: a maximum of 100 attempts, a depth limit of 5 reasoning steps, and token budgets of 400 per function (100 for OA). Unsolved problems are retried twice before being discarded. This approach combines the heuristic power of A* with computational practicality for structured reasoning.

\subsection{Pseudocode-guided Reasoning}

To enable structured and reliable reasoning for robotic automation systems, we represent each reasoning trajectory as a sequence of executable abstract functions expressed in pseudocode. This modular representation ensures that VLMs perform step-by-step inference consistent with logically decomposable task structures, significantly mitigating hallucinations and error propagation.
To adapt these pseudocode paths to a specific question, we introduce a hybrid selection strategy that considers both question difficulty and semantic content. This ensures that the retrieved paths are not only contextually relevant but also cognitively appropriate. Formally, the Hybrid Similarity Score (HSS) used to retrieve reasoning paths is defined as:

\begin{align}
\label{eq:hss}
\nonumber
\text{HSS} =& \alpha \cdot \| \text{DFV}_k - 
\text{DFV}_i \|_2\\&-(1-\alpha) \cdot \cos(\text{emb}_{Q_k}, \text{emb}_{Q_i})
\end{align}

\definecolor{mygreen}{RGB}{0,150,0}
\definecolor{myred}{RGB}{150,0,0}

\begin{table*}[t]
\centering
\setlength{\tabcolsep}{3pt}
\renewcommand{\arraystretch}{1.4}
\resizebox{\textwidth}{!}{
    \begin{tabular}{l|cccc|cccc|ccccccc|l}
    \toprule
    \textbf{Model} & \multicolumn{4}{c|}{\textbf{POPE}} & \multicolumn{4}{c|}{\textbf{HallusionBench}} & \multicolumn{7}{c|}{\textbf{MMStar}} & \textbf{AVG}\\
     & Acc & Prec & Rec & Overall & aAcc & fAcc & qAcc & Overall & CP & FP & IR & LR & MATH & ST & Overall & \\
    \midrule
    InternVL2-2B* & 86.7 & 95.7 & 76.8 & 85.2 & 58.7 & 26.0 & 29.2 & 38.0 & 64.0 & 38.8 & 60.8 & 42.8 & 49.2 & 43.2 & 49.8 & 57.7 \\
    InternVL2-8B* & 86.0 & 96.6 & 74.7 & 84.2 & 63.9 & 35.0 & 36.0 & 45.0 & 70.4 & 48.8 & 68.0 & 63.2 & 67.2 & 51.6 & 61.5 & 63.6 \\
    LLaMA-3.2-11B-Vision* & 88.1 & 88.0 & \textbf{88.3} & 88.1 & 58.0 & 31.8 & 31.2 & 40.3 & 66.0 & 46.4 & 57.6 & 50.8 & 45.2 & 32.8 & 49.8 & 59.4 \\
    LLaVA-v1.5-7B* & 87.0 & 92.1 & 80.9 & 86.1 & 48.8 & 20.5 & 13.6 & 27.6 & 57.2 & 24.4 & 41.6 & 28.4 & 26.4 & 20.4 & 33.1 & 48.9 \\
    GPT-4V (0409)* & 83.9 & 93.9 & 72.5 & 81.8 & 62.1 & 36.7 & 32.7 & 43.9 & 69.6 & 44.8 & 67.2 & 59.2 & 60.8 & 34.4 & 56.0 & 60.6 \\
    Qwen2-VL-2B* & 88.2 & 94.7 & 81.0 & 83.7 & 61.5 & 32.4 & 33.0 & 42.4 & 60.0 & 50.0 & 58.4 & 43.2 & 40.8 & 32.8 & 47.5 & 59.5 \\
    Qwen2-VL-7B* & 89.1 & 94.9 & 82.7 & 88.4 & 68.5 & 39.0 & 43.7 & 50.4 & 72.0 & 56.4 & 68.0 & 61.6 & 58.4 & 48.0 & 60.7 & 66.5 \\
    Qwen2.5-VL-7B* & 87.4 & 96.8 & 77.3 & 85.9 & 69.9 & \textbf{41.3} & \textbf{44.4} & 51.9 & 72.0 & 57.6 & 70.4 & 67.6 & 67.2 & 49.6 & 64.1 & 67.3 \\
    \hline
    \textbf{Ours} (Qwen2-VL-2B) & \textbf{88.7} &	98.1&	78.8&	88.5&	75.7&	23.0&	31.7&	43.4&	55.6&	35.6&	61.6&	54.4&	56.4&	38.8&	50.4&	60.8$^{\textcolor{mygreen}{+1.7}}$
 \\
    \textbf{Ours} (Qwen2-VL-7B) & \textbf{88.7}&	\textbf{99.9}&	77.4&	\textbf{88.7}&	\textbf{81.8}&	33.6&	42.8&	52.7&	74.8&	60.0&	69.6&	66.8&	67.6&	\textbf{55.6}&	65.7&	69.0$^{\textcolor{mygreen}{+2.5}}$
 \\
    \textbf{Ours} (Qwen2.5-VL-7B) & 87.3&	97.7&	76.3&	87.1&	81.4&	36.1&	41.1&	\textbf{52.9}&	\textbf{76.8}&	\textbf{61.2}&	\textbf{70.8}&	\textbf{73.2}&	\textbf{72.4}&	53.6&	\textbf{68.0}&	\textbf{69.3}$^{\textcolor{mygreen}{+2.0}}$
 \\
    \bottomrule
    \end{tabular}
}
% textbf{64.4}$^{\textcolor{mygreen}{+5.0}}$
\caption{Comparison of VLMs across POPE, HallusionBench, and MMStar benchmarks. The AVG score represents the average of the overall scores across the three benchmarks.
\textbf{Bold} values indicate the highest score in each column, and \textcolor{mygreen}{green} values show the improvement over the corresponding backbone model.
All values are presented as percentages. 
Models marked with an asterisk (*) indicate that the data is sourced from the OpenVLM Leaderboard \cite{VLMEvalkit}.
}
\label{tab:multimodal_comparison}
\end{table*}

\noindent The first term measures L2 distance between the Difficulty Feature Vectors (DFVs), capturing differences in reasoning complexity. The second term computes cosine similarity between question embeddings to ensure semantic alignment. The hyperparameter \( \alpha \in [0,1] \) controls the trade-off between difficulty and semantic relevance.
By combining this hybrid retrieval strategy with Pseudocode-guided reasoning, PStar selects well-matched, interpretable reasoning paths that constrain the model’s behavior, reduce hallucinations, and improve answer reliability across diverse multimodal reasoning tasks.

\section{Experiments}

\subsection{Experimental Settings}
\label{sec:settings}
To evaluate the applicability of our method in realistic automation scenarios, we selected lightweight yet capable VLMs as our base systems: \textbf{Qwen2-VL-2B-Instruct}, \textbf{Qwen2-VL-7B-Instruct}, and \textbf{Qwen2.5-VL-7B-Instruct}. These models represent a practical balance between computational efficiency and reasoning performance—a critical consideration for deployment in resource-constrained robotic platforms requiring real-time visual-language understanding.

Inference uses bfloat16 in PyTorch. We set $M = 3000$ (Eq.~\ref{eq:f_s_all}), $\alpha = 0.5$ (Eq.~\ref{eq:hss}), cost coefficients $\lambda_{RR}=1.0$, $\lambda_{SR,ER}=1.8$, and $\lambda_{\ne(RR,SR,ER)}=1.6$. Token limits are 400 for abstract functions (except OA=100). Pseudocode library construction used Qwen2.5-VL-7B-Instruct (repetition penalty: 1.05, temperature: 1.0, top-$p$: 0.9); evaluation used repetition penalty: 1.05, temperature: 0.5, top-$p$: 0.9. Experiments leverage MS-Swift~\cite{MS-Swift}.

\subsection{Benchmarks}
\label{sec:benchmarks}

We evaluate our method on benchmarks addressing open-world robotic challenges in perception and reasoning. Key metrics include hallucination resistance, spatial understanding, and logical inference.
\textbf{POPE} \cite{POPE} tests object hallucination robustness using accuracy/precision/recall metrics, relevant for manipulation tasks.
\textbf{HallusionBench} \cite{HallusionBench} evaluates fine-grained reasoning through aAcc (average accuracy), fAcc (accuracy per image across all questions), and qAcc (accuracy per question across all images).
\textbf{MMStar} \cite{MMStar} assesses six capabilities(coarse perception, fine-grained perception, instance reasoning, logical reasoning, math, and science \& technology), covering essential robotic automation demands.

\subsection{Baselines}
We evaluate our approach against a diverse set of multimodal baselines, including \textbf{InternVL2-2B} and \textbf{InternVL2-8B} \cite{InternVL2-0}, a series of VLMs designed for comprehensive multimodal understanding. We also compare with \textbf{Llama-3.2-11B-Vision-Instruct} \cite{LLama3}, an instruction-tuned LLM with integrated vision capabilities. Additionally, we include \textbf{LLaVA-v1.5-7B} \cite{LLAVA1-5}, which enhances multimodal understanding by combining CLIP-based vision encoding with large language models. Finally, \textbf{GPT-4V} (0409, detail-high) \cite{GPT-4V} serves as a proprietary SOTA baseline for vision-language reasoning.

\subsection{Experimental Results}
As shown in Table~\ref{tab:multimodal_comparison}, our method Pstar consistently improves the performance of various base models across all three benchmarks.
\begin{itemize}
\item \textbf{POPE:} PStar consistently improves factuality-related metrics across different backbones. Notably, built on Qwen2-VL-2B, PStar achieves the overall score of 88.5 with a precision of 98.1 and a recall of 78.8, outperforming the base model by 4.8. With Qwen2.5-VL-7B, PStar lifts the overall score from 85.9 to 87.1.
    
\item \textbf{HallusionBench:} Our model also yields consistent improvements in HallusionBench. For instance, on Qwen2.5-VL-7B, it raises the overall score from 51.9 to 52.9, surpassing larger models such as GPT-4V. Significant gains are also observed on Qwen2-VL-7B (+2.3), highlighting the effectiveness of our approach across model scales.
    
\item \textbf{MMStar:} PStar brings notable improvements across six capabilities. Using Qwen2.5-VL-7B, we achieve the highest scores on coarse perception (76.8), fine-grained perception (61.2), and instance reasoning (70.8), leading to a new state-of-the-art overall score of 68.0.
\end{itemize}
On average across the three benchmarks, PStar achieves the highest score of 69.3 when built upon Qwen2.5-VL-7B, demonstrating its effectiveness in enhancing both general perception and fine-grained reasoning.

\begin{table*}[h]
\centering
\setlength{\tabcolsep}{3pt}
\renewcommand{\arraystretch}{1.15}
\resizebox{0.9\textwidth}{!}{
    \begin{tabular}{l|cccc|ccccccc|l}
    \toprule
    \textbf{Method} & \multicolumn{4}{c|}{\textbf{HallusionBench}} & \multicolumn{7}{c|}{\textbf{MMStar}} & AVG \\
      & aAcc & fAcc & qAcc & Overall & CP & FP & IR & LR & MATH & ST & Overall & \\
    \midrule
Qwen2-VL-2B & 61.8&	32.4&	33.0&	42.4&	60.0&	50.0&	58.4&	43.2&	40.8&	32.8&	47.5&	45.0
\\
+ SFT (RLAIF, 80k) & 51.3 & 21.4 & 21.3 & 31.3 & 44.8 & 26.8 & 37.2 & 25.6 & 26.0 & 12.8 & 28.9 & 30.1 \\
+ CPO (RLAIF, 10k) & 47.4 & 19.4 & 21.5 & 29.4 & 40.0 & 23.6 & 39.6 & 27.2 & 31.2 & 20.4 & 30.3 & 29.9 \\
+ SFT (Mixed, 6k) & 35.0 & 5.8 & 11.2 & 17.3 & 49.6 & 32.4 & 38.0 & 32.0 & 40.4 & 15.6 & 34.7 & 26.0 \\
+ CPO (Mixed, 6k) & 35.8 & 4.0 & 10.8 & 16.9 & 53.2 & 31.2 & 39.6 & 33.6 & 37.2 & 13.6 & 34.7 & 25.8 \\
\rowcolor{gray!15}
Ours (Qwen2-VL-2B) & 75.7&	23.0&	31.7&	43.4&	55.6&	35.6&	61.6&	54.4&	56.4&	38.8&	50.4&	46.9
 \\
\hline
Qwen2-VL-7B & 68.5&	39.0&	43.7&	50.4&	72.0&	56.4&	68.0&	61.6&	58.4&	48.0&	60.7&	55.6
\\
+ SFT (RLAIF, 80k) & 56.8 & 28.0 & 28.4 & 37.7 & 42.4 & 22.8 & 42.8 & 34.8 & 32.8 & 19.2 & 32.5 & 35.1 \\
+ CPO (RLAIF, 10k) & 61.4 & 36.1 & 38.2 & 45.3 & 45.2 & 26.0 & 43.6 & 35.6 & 44.0 & 24.8 & 36.5 & 40.9 \\
+ SFT (Mixed, 6k) & 41.1 & 7.2 & 15.4 & 21.2 & 48.4 & 37.6 & 44.4 & 47.2 & 53.2 & 18.8 & 41.6 & 31.4 \\
+ CPO (Mixed, 6k) & 39.7 & 4.6 & 13.2 & 19.2 & 52.0 & 41.2 & 46.8 & 43.6 & 47.6 & 16.8 & 41.3 & 30.3 \\
\rowcolor{gray!15}
Ours (Qwen2-VL-7B)& 81.8&	33.6&	42.8&	52.7&	74.8&	60.0&	69.6&	66.8&	67.6&	55.6&	65.7&	59.2
 \\
\hline
\end{tabular}
}
\vspace{1ex}
\caption{Comparison of training strategies on HallusionBench and MMStar.}
\label{tab:comparison}
\end{table*}

\newcommand{\cmark}{\textcolor{green!60!black}{\checkmark}} % green checkmark
\newcommand{\xmark}{\textcolor{red!70!black}{×}}    % red cross

\begin{table}[h]
\setlength{\tabcolsep}{3pt}
\resizebox{\columnwidth}{!}{
    \begin{tabular}{lcccc}
    \toprule
    \textbf{Method} & \textbf{MMStar} & \textbf{Open-Source} & \textbf{Data} & \textbf{Training-Free} \\
    \midrule
    Mulberry \cite{Mulberry}     & 61.3\% & \xmark & 260k  & \xmark \\
    AStar \cite{AStar}       & 61.7\% & \cmark & 0.5k  & \cmark \\
    LLaVA-CoT \cite{LLAVA-COT}   & 57.6\% & \xmark & 100k  & \xmark \\
    LlamaV-o1 \cite{LLamaV-o1}   & 59.5\% & \xmark & 118k  & \xmark \\
    \rowcolor{gray!15}
    \textbf{Ours} & \textbf{68.0\%} & \cmark & 0.5k  & \cmark \\
    \bottomrule
    \end{tabular}
}
\vspace{1ex}
\caption{Comparison of methods on MMStar benchmark.}
\label{tab:mmstar_comparison}
\end{table}

\begin{table}[t]
\centering
\begin{tabular}{l|l}
\toprule
\multicolumn{2}{c}{\textbf{Mixed [MCQA 1k]}} \\
\midrule
\textbf{Reasoning Path} & \textbf{Accuracy} \\
\midrule
Vanilla & 51.0 \\
SA() RR() RR() & 59.3$^{\textcolor{mygreen}{+8.3}}$ \\
SA() RR() VA() RR() & 58.8$^{\textcolor{mygreen}{+7.8}}$ \\
SA() VA() NA() RR() SP() RR() SR() RR() & 58.9$^{\textcolor{mygreen}{+7.9}}$ \\
SA() VA() NA() SP() RR() RR() SR() RR() & 60.1$^{\textcolor{mygreen}{+9.1}}$ \\
\midrule
\multicolumn{2}{c}{\textbf{MathVerse TestMini [MCQA 1k]}} \\
\midrule
\textbf{Reasoning Path} & \textbf{Accuracy} \\
\midrule
Vanilla & 40.5 \\
SA() RR() RR() & 32.3$^{\textcolor{myred}{-8.2}}$ \\
SA() RR() VA() RR() & 41.8$^{\textcolor{mygreen}{+1.3}}$ \\
SR() VA() NA() RR() SP() RR() SR() RR() & 42.4$^{\textcolor{mygreen}{+1.9}}$ \\
SA() VA() NA() SP() RR() RR() SR() RR() & 44.7$^{\textcolor{mygreen}{+4.2}}$ \\
\midrule
\multicolumn{2}{c}{\textbf{Mixed [OEQA 1k]}} \\
\midrule
\textbf{Reasoning Path} & \textbf{Accuracy} \\
\midrule
Vanilla & 12.7 \\
SA() RR() RR() & 15.8$^{\textcolor{mygreen}{+3.1}}$ \\
SA() VA() NA() SP() RR() RR() SR() RR() & 15.7$^{\textcolor{mygreen}{+3.0}}$ \\
\midrule
\multicolumn{2}{c}{\textbf{Mixed [YN 364]}} \\
\midrule
\textbf{Reasoning Path} & \textbf{Accuracy} \\
\midrule
Vanilla & 43.1 \\
SA() RR() RR() & 47.5$^{\textcolor{mygreen}{+4.4}}$ \\
SA() VA() NA() SP() RR() RR() SR() RR() & 35.4$^{\textcolor{myred}{-7.7}}$ \\
\bottomrule
\end{tabular}

\vspace{1ex}
\caption{Performance of \textbf{Qwen2-VL-7B} under different fixed reasoning paths across multiple VQA benchmarks. Vanilla means no Pseudocode-guided reasoning.}
\label{tab:reasoning_path_results_split}
\end{table}

\begin{table}[t]
\centering

\resizebox{0.8\columnwidth}{!}{
\begin{tabular}{l|c|c}
\toprule
\textbf{Type} & \textbf{Number} & \textbf{Ratio (\%)} \\
\midrule
Correct then Correct & 4370 & 42.60 \\
Correct then Wrong   & 647  & 6.31 \\
Wrong then Correct   & 1540 & 15.01 \\
Wrong then Wrong     & 3701 & 36.08 \\
\bottomrule
\end{tabular}
}
\vspace{0.5ex}
\caption{Consistency outcomes across two rounds of reasoning on 10,258 Mixed samples.}
\label{tab:answer_transition}
\end{table}

\begin{table}[t]
\resizebox{\columnwidth}{!}{
    \begin{tabular}{l|ccccccc}
    \toprule
    \textbf{Model} & \multicolumn{7}{c}{\textbf{MMStar}}\\
     & CP & FP & IR & LR & MATH & ST & Overall\\
    \midrule
    \rowcolor{gray!15}
    Ours & 76.8 &61.2&70.8&73.2&72.4&53.6&68.0\\
    w/o hybrid search & 70.0 & 48.8 & 60.4 & 55.2 & 59.2 & 39.6 & 55.5\\
    w/o DFV & 70.8 & 45.2 & 64.0 & 51.6 & 56.4 & 40.8 & 54.8  \\
    \bottomrule
    \end{tabular}
}
\vspace{0.5ex}
\caption{Ablation study on MMStar. }
\label{tab:ablation}
\end{table}

\subsection{Comparative Performance Analysis}
\paragraph{Comparison with Existing Methods}
To further highlight the advantages of our method, we present a quantitative comparison against representative baselines in Table~\ref{tab:mmstar_comparison}, focusing on four dimensions: MMStar accuracy, open-source availability, data volume, and training-free design.
Our method achieves the highest accuracy (68.0\%), outperforming previous tree-based reasoning approaches such as Mulberry (61.3\%) and AStar (61.7\%).
Notably, while methods like Mulberry and LLaVA-CoT rely on over 100k training samples, our approach reaches state-of-the-art performance with only 0.5k examples, demonstrating strong data efficiency.
In addition, our method is entirely built on open-source tools and models, ensuring broad reproducibility and transparency.
Overall, this comparison shows that our approach not only advances multimodal reasoning performance but also sets a new standard in efficiency and accessibility.

\paragraph{Comparison with Training Methods}
We compare our training-free method with two popular paradigms: supervised fine-tuning (SFT) and contrastive preference optimization (CPO) \cite{CPO}. SFT uses reference answers for supervision, while CPO optimizes preferences through contrastive learning. We evaluate on two datasets: \textbf{RLAIF-V} \cite{RLAIF-V}, with GPT-4V-generated answers; and \textbf{Mixed} dataset (MathVista \cite{MathVista}, MATH-Vision \cite{MathVision}, and ScienceQA \cite{ScienceQA}).

As shown in Table~\ref{tab:comparison}, our method consistently outperforms SFT and CPO, with a +1.9\% gain in the 2B setting. 
While training generally enhances model capabilities, performance degradation occurs due to the domain gap between training data and evaluation benchmarks - for instance, when models trained on open-ended RLAIF data are evaluated on multiple-choice tasks like HallusionBench and MMStar, or when mathematical reasoning tasks from Mixed dataset introduce distributional shifts.
In contrast, our method avoids this and achieves the highest MMStar score (65.7\%) at 7B, with a +5.0\% gain. These results demonstrate the scalability, robustness, and efficiency of our training-free approach.

\subsection{Evaluation of Fixed Pseudocode-guided Reasoning Paths}
We evaluated the impact of fixed reasoning paths on model performance using Qwen2-VL-7B, focusing on whether structured reasoning logic improves accuracy in visual question answering (VQA) tasks. The evaluation was conducted on the \textbf{Mixed dataset} (MathVista \cite{MathVista}, Math-Vision \cite{MathVision}, and ScienceQA \cite{ScienceQA}), with 1k multiple-choice (MCQA), 1k open-ended (OEQA), and 364 yes/no (YN) questions, as well as the MathVerse TestMini set (1k MCQA samples).

By comparing predefined reasoning paths with the Vanilla, we found that Pseudocode-guided paths improved accuracy. For instance, the best path for Mixed [MCQA] showed a 9.1-point improvement, and gains of 3.1 and 4.4 points were observed on Mixed [OEQA] and [YN], respectively. While performance varied across datasets, the overall trend suggests that structured reasoning paths can enhance model reliability and reduce hallucinations, making them a promising strategy for improving multimodal models.

\subsection{Consistency Analysis}

We investigated the consistency and reliability of our model's reasoning process by evaluating its performance on repeated executions of abstract functions. Specifically, we designed a fixed reasoning path \texttt{RR() OA() SR() RR() OA()} to simulate a two-round answering process, where the model reflects on and re-executes reasoning to generate a final answer. We applied this process to all 10{,}258 examples from the \textbf{Mixed} dataset.

The results, summarized in Table~\ref{tab:answer_transition}, showed that 42.60\% of answers were consistent in both rounds. Notably, 15.01\% of answers improved from wrong to correct in the second round, highlighting the benefit of self-reflection. However, 6.31\% regressed from correct to wrong, and 36.08\% remained incorrect in both rounds. These findings suggest that self-reflection enhances model robustness by encouraging consistency and reducing hallucinations.

\begin{figure}[t]
  \centering
  \begin{subfigure}[b]{0.23\textwidth}
    \includegraphics[width=\textwidth]{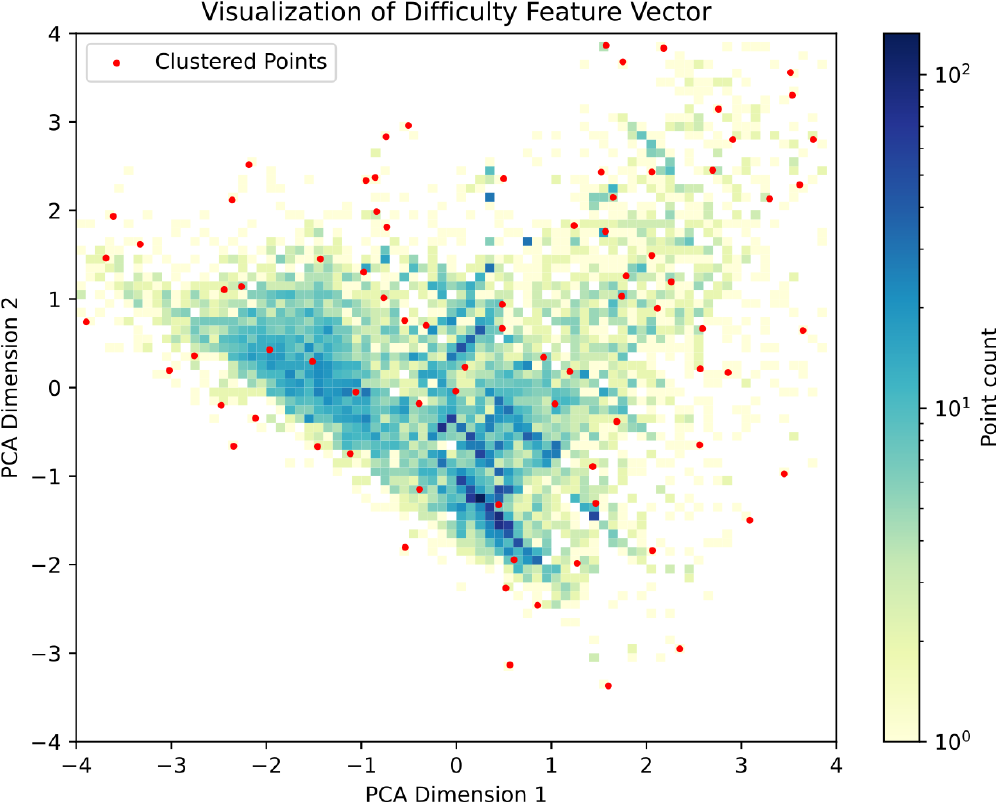}
    % \caption{\(k = 100\)}
  \end{subfigure}
  \begin{subfigure}[b]{0.23\textwidth}
    \includegraphics[width=\textwidth]{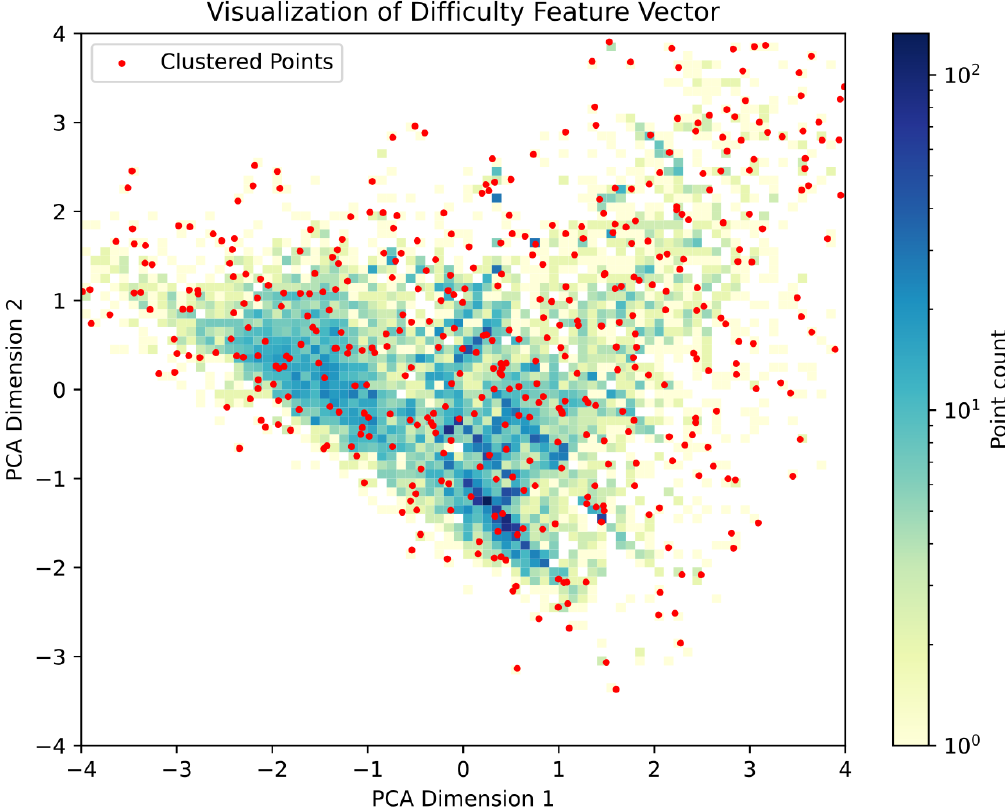}
    % \caption{\(k = 500\)}
  \end{subfigure}
  \caption{Visualization of DFVs in PCA space for seed sizes \(k = 100\) (left) and \(k = 500\) (right). Red points indicate max-min distance-selected seeds overlaid on a density heatmap. }
  \label{fig:two_images}
\end{figure}

\begin{figure*}[t]
    \centering
    \includegraphics[width=1.0\linewidth]{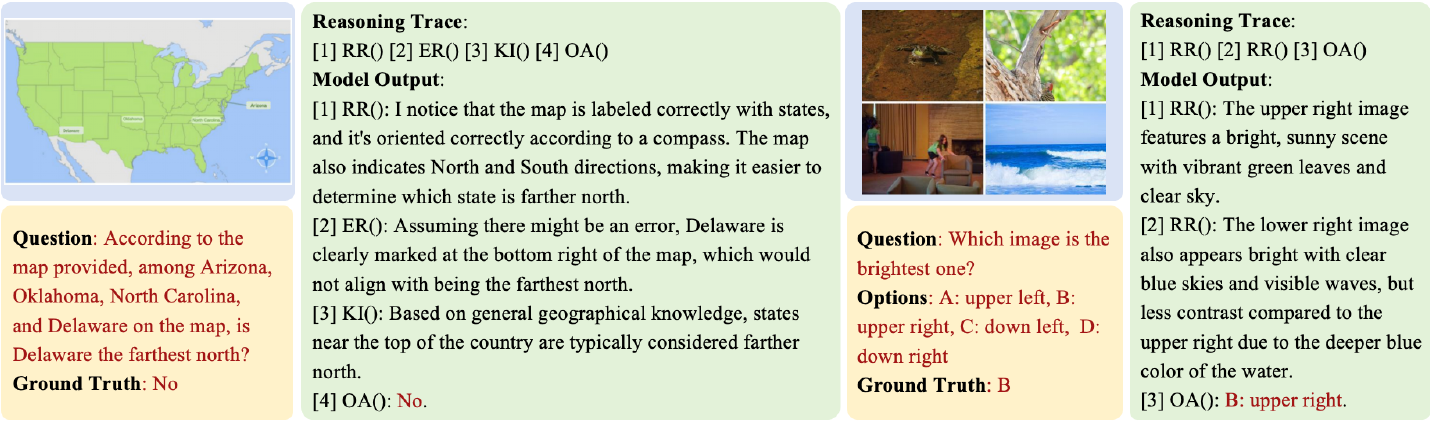}
    \caption{Use case study of two examples.}
    \label{fig:use_case}
\end{figure*}

\begin{table}[t]
\resizebox{\columnwidth}{!}{
    \begin{tabular}{l|ll}
    \toprule
    \textbf{Model} & \multicolumn{2}{c}{\textbf{Accuracy}}\\
     & LVM Evaluation & Strict Match\\
    \midrule
    Deepseek-vl-7B & 56.9 & 42.9 \\
    \rowcolor{gray!15}
    Ours (Deepseek-vl-7B) & 60.2$^{\textcolor{mygreen}{+3.3}}$ & 47.0$^{\textcolor{mygreen}{+4.1}}$ \\
    Qwen2-VL-7B & 70.8 & 55.7 \\
    \rowcolor{gray!15}
    Ours (Qwen2-VL-7B) & 71.6$^{\textcolor{mygreen}{+0.8}}$ & 56.7$^{\textcolor{mygreen}{+1.0}}$ \\
    Qwen2.5-VL-7B & 69.5 & 50.0 \\
    \rowcolor{gray!15}
    Ours (Qwen2.5-VL-7B) & 70.5$^{\textcolor{mygreen}{+1.0}}$ & 55.0$^{\textcolor{mygreen}{+5.0}}$ \\
    \bottomrule
    \end{tabular}
}
\vspace{0.5ex}
\caption{Comparisons of general open-ended tasks on OKVQA. }
\label{tab:okvqa}
\end{table}

\subsection{Ablation Study and Analysis}

\paragraph{Ablation Study}
We ablate two key components: hybrid search and DFV. Replacing guided paths with random ones leads to a 13.2 point drop with Qwen2.5-VL-7B, showing the value of structured path selection. Removing DFV and using only question embeddings causes a similar 12.5 point drop, highlighting the role of multimodal features. 
% Table~\ref{tab:multimodal_comparison} shows that omitting the pseudocode framework or abstract functions degrades reasoning. 
% Full results are in Appendix~\ref{sec:appendix_ablation_study}.
% We ablate two key components: hybrid search and DFV. Removing hybrid search by replacing guided paths with random ones consistently reduces performance, causing an average drop of 5.6 points with Qwen2.5-VL-7B, which demonstrates the value of structured path selection. Excluding DFV and relying solely on question embeddings also hurts performance, resulting in a similar average decline of 5.7 points, highlighting the importance of multimodal features. Previous experiments shown in Table~\ref{tab:multimodal_comparison} further confirm that removing the pseudocode framework or abstract functions impairs reasoning ability, reinforcing the benefit of our structured approach. Full results can be found in Appendix~\ref{sec:appendix_ablation_study}.

\paragraph{DFV Visualization}
We visualize DFVs from multitask datasets (projected via PCA) to assess seed coverage. With sample sizes \(k = 100\) and \(k = 500\), heatmaps reveal a dense core and sparse periphery. Red dots (selected by max-min distance) are well spread, covering diverse difficulty regions. Larger \(k\) improves representativeness and ensures broader spectrum coverage, which in turn enhances seed quality and contributes to more robust reasoning performance.
 (Figure~\ref{fig:two_images}).

\paragraph{Use Case Study}
Figure~\ref{fig:use_case} presents two representative multimodal questions along with their corresponding reasoning paths generated by PStar. The model follows these Pseudocode-guided paths step by step, executing each abstract function to perform structured reasoning. This example illustrates how PStar enables interpretable, accurate, and hallucination-resistant answer generation by aligning reasoning strategies with question complexity.

\paragraph{Open-Ended Reasoning for Embodied Tasks}
To evaluate the general robustness and task-oriented reasoning capability of our method in open-world settings, we benchmarked it on the OKVQA dataset \cite{OKVQA}, which requires contextual and commonsense reasoning similar to that needed for robotic instruction following and human-robot interaction. We used Qwen2.5-VL-7B as the base model with a Strict Match metric. As shown in Table~\ref{tab:okvqa}, our approach consistently improves performance across all base models. Notably, it achieves a significant +5.0\% improvement in Strict Match accuracy for Qwen2.5-VL-7B. These results confirm that our method enhances the reliability and general applicability of visual question answering—a capability critical for robots operating in unstructured environments where interpretative robustness directly impacts task success.

\section{Related Work}
Vision-Language Models (VLMs) have become integral to robotic systems, enabling advanced capabilities such as visual instruction following \cite{robot_vqa_surgery} and human-object interaction reasoning \cite{automatic_HOI}. However, their integration into safety-critical autonomous applications is severely hindered by the persistent challenge of \textit{hallucination}—the generation of textual or decisional outputs that are inconsistent with visual inputs \cite{MiniGPT-4}. Such errors can lead to catastrophic failures in real-world deployments, where incorrect perception or reasoning directly impacts physical actions.
Recent studies have begun to systematically categorize the origins of VLM hallucination, attributing it to four key aspects: data imperfections, model architectures, training strategies, and inference mechanisms \cite{bai2025hallucinationmultimodallargelanguage}.
Inadequate training data may hinder effective alignment between modalities, the quality and static bias of training data also pose significant obstacles.
Prevailing VLMs architechture consists of a vision encoder, a projector for cross-modal alignment, and a pretrained LLM. Suboptimal or erroneous outputs from each module may lead to hallucinations ~\cite{HallusionBench}.
Regarding training and inference, \cite{2024rlhfv} suggest that lack of RLHF stage becomes a potention cause of hallucinations, while several other studies argue that attention to visual content becomes increasingly diluted as the sequence length grows during generation\cite{huang2024opera}.

\subsection{Reliable Automation of VLMs}

\textbf{Conventional Hallucination Mitigation Strategies} Current research on hallucination mitigation of VLMs focuses on external tool augmentation, training-based methods and decoding strategies. 
Woodpecker~\cite{yin2024woodpecker} combines VLMs and a pretrained VQA model to detect hallucinations in outputs and correct errors.
Training-based methods involve crafting instruction-tuning datasets, such as MedHallTune ~\cite{yan2025medhalltuneinstructiontuningbenchmarkmitigating}, or enhancing cross-modal alignment through alignment training ~\cite{Xiao_Huang_Gan_He_Li_Yu_Shu_Jiang_Zhu_2025}. 
However, due to high labor and computational costs, training-free decoding methods have been studied. OPERA \cite{huang2024opera} introduces a penalty term along with a rollback strategy to prevent overtrust on summary tokens. 

\textbf{Reasoning-Based Approaches}
Incorporating Chain-of-Thought into VLMs has led to notable improvements.
MMCoT~\cite{zhang2023multicot} finetunes language models to perform Multimodal CoT, revealing its potential in mitigating hallucinations and enhancing convergence speed. 
KAM-CoT~\cite{mondal2024kamcot}, which utilizes multimodal CoT reasoning with knowledge graphs, helps the model to gain a deeper contextual understanding, thus reducing hallucinations and enhancing the quality of answers.

\section{Conclusion And Future Work}
We presented \textbf{PStar}, a Pseudocode-guided structured reasoning framework designed to enhance the reliability and safety of vision-language models (VLMs) in robotic automation. By adaptively selecting reasoning paths through a library of abstract functions and a Difficulty Feature Vector, PStar reduces hallucinations and supports interpretable, step-by-step reasoning essential for real-world robotic tasks such as instruction following, scene understanding, and action planning. Extensive experiments demonstrate that PStar significantly improves robustness and accuracy, achieving state-of-the-art performance on benchmarks including POPE (87.1) and MMStar (68.0), indicating strong potential for deployment in embodied AI systems.

While our method reduces computational overhead, the offline A* search process remains computationally expensive. We aim to expand into open-world tasks such as human-robot collaboration and long-horizon planning, thereby enhancing adaptive behavior in unstructured environments.



% \section*{ACKNOWLEDGMENT}

% The preferred spelling of the word ÒacknowledgmentÓ in America is without an ÒeÓ after the ÒgÓ. Avoid the stilted expression, ÒOne of us (R. B. G.) thanks . . .Ó  Instead, try ÒR. B. G. thanksÓ. Put sponsor acknowledgments in the unnumbered footnote on the first page.

%%%%%%%%%%%%%%%%%%%%%%%%%%%%%%%%%%%%%%%%%%%%%%%%%%%%%%%%%%%%%%%%%%%%%%%%%%%%%%%%

% References are important to the reader; therefore, each citation must be complete and correct. If at all possible, references should be commonly available publications.


\begin{thebibliography}{99}

\bibitem{LLAVA1-5}
H.~Liu, C.~Li, Y.~Li, and Y.~J. Lee, ``Improved baselines with visual instruction tuning,'' in \emph{Proceedings of the IEEE/CVF Conference on Computer Vision and Pattern Recognition}, 2024, pp. 26\,296--26\,306.

\bibitem{GPT-4V}
Z.~Yang, L.~Li, K.~Lin, J.~Wang, C.-C. Lin, Z.~Liu, and L.~Wang, ``The dawn of lmms: Preliminary explorations with gpt-4v (ision),'' \emph{arXiv preprint arXiv:2309.17421}, vol.~9, no.~1, p.~1, 2023.

\bibitem{MiniGPT-4}
D.~Zhu, J.~Chen, X.~Shen, X.~Li, and M.~Elhoseiny, ``Minigpt-4: Enhancing vision-language understanding with advanced large language models,'' in \emph{ICLR}, 2024.

\bibitem{InternVL2-0}
Z.~Chen, W.~Wang, Y.~Cao, Y.~Liu, Z.~Gao, E.~Cui, J.~Zhu, S.~Ye, H.~Tian, Z.~Liu \emph{et~al.}, ``Expanding performance boundaries of open-source multimodal models with model, data, and test-time scaling,'' \emph{arXiv preprint arXiv:2412.05271}, 2024.

\bibitem{RLAIF-V}
T.~Yu, H.~Zhang, Q.~Li, Q.~Xu, Y.~Yao, D.~Chen, X.~Lu, G.~Cui, Y.~Dang, T.~He, X.~Feng, J.~Song, B.~Zheng, Z.~Liu, T.-S. Chua, and M.~Sun, ``Rlaif-v: Open-source ai feedback leads to super gpt-4v trustworthiness,'' 2024. 

\bibitem{LLAVA-COT}
G.~Xu, P.~Jin, H.~Li, Y.~Song, L.~Sun, and L.~Yuan, ``Llava-cot: Let vision language models reason step-by-step,'' 2025.

\bibitem{CPO}
H.~Xu, A.~Sharaf, Y.~Chen, W.~Tan, L.~Shen, B.~Van~Durme, K.~Murray, and Y.~J. Kim, ``Contrastive preference optimization: Pushing the boundaries of llm performance in machine translation,'' in \emph{International Conference on Machine Learning}.\hskip 1em plus 0.5em minus 0.4em\relax PMLR, 2024, pp. 55\,204--55\,224.


\bibitem{POPE}
Y.~Li, Y.~Du, K.~Zhou, J.~Wang, W.~X. Zhao, and J.-R. Wen, ``Evaluating object hallucination in large vision-language models,'' in \emph{Proceedings of the 2023 Conference on Empirical Methods in Natural Language Processing}, 2023, pp. 292--305.

\bibitem{HallusionBench}
T.~Guan, F.~Liu, X.~Wu, R.~Xian, Z.~Li, X.~Liu, X.~Wang, L.~Chen, F.~Huang, Y.~Yacoob \emph{et~al.}, ``Hallusionbench: an advanced diagnostic suite for entangled language hallucination and visual illusion in large vision-language models,'' in \emph{Proceedings of the IEEE/CVF Conference on Computer Vision and Pattern Recognition}, 2024, pp. 14\,375--14\,385.

\bibitem{MMStar}
L.~Chen, J.~Li, X.~Dong, P.~Zhang, Y.~Zang, Z.~Chen, H.~Duan, J.~Wang, Y.~Qiao, D.~Lin \emph{et~al.}, ``Are we on the right way for evaluating large vision-language models?'' \emph{arXiv preprint arXiv:2403.20330}, 2024.

\bibitem{MathVista}
P.~Lu, H.~Bansal, T.~Xia, J.~Liu, C.~Li, H.~Hajishirzi, H.~Cheng, K.-W. Chang, M.~Galley, and J.~Gao, ``Mathvista: Evaluating mathematical reasoning of foundation models in visual contexts,'' in \emph{International Conference on Learning Representations (ICLR)}, 2024.

\bibitem{MathVision}
K.~Wang, J.~Pan, W.~Shi, Z.~Lu, H.~Ren, A.~Zhou, M.~Zhan, and H.~Li, ``Measuring multimodal mathematical reasoning with math-vision dataset,'' in \emph{The Thirty-eight Conference on Neural Information Processing Systems Datasets and Benchmarks Track}, 2024. 

\bibitem{ScienceQA}
P.~Lu, S.~Mishra, T.~Xia, L.~Qiu, K.-W. Chang, S.-C. Zhu, O.~Tafjord, P.~Clark, and A.~Kalyan, ``Learn to explain: Multimodal reasoning via thought chains for science question answering,'' in \emph{The 36th Conference on Neural Information Processing Systems (NeurIPS)}, 2022.

\bibitem{Qwen2.5-VL}
Q.~Team, ``Qwen2.5-vl,'' January 2025.

\bibitem{Hallucination_Snowball}
X.~Liang, S.~Song, Z.~Zheng, H.~Wang, Q.~Yu, X.~Li, R.-H. Li, F.~Xiong, and Z.~Li, ``Internal consistency and self-feedback in large language models: A survey,'' \emph{CoRR}, 2024.

\bibitem{LLama3}
A.~Grattafiori and T.~Abhimanyu~Dubey, ``The llama 3 herd of models,'' 2024.

\bibitem{MS-Swift}
Y.~Zhao, J.~Huang, J.~Hu, X.~Wang, Y.~Mao, D.~Zhang, Z.~Jiang, Z.~Wu, B.~Ai, A.~Wang \emph{et~al.}, ``Swift: a scalable lightweight infrastructure for fine-tuning,'' in \emph{Proceedings of the AAAI Conference on Artificial Intelligence}, vol.~39, no.~28, 2025, pp. 29\,733--29\,735.

\bibitem{Underthinking}
Y.~Wang, Q.~Liu, J.~Xu, T.~Liang, X.~Chen, Z.~He, L.~Song, D.~Yu, J.~Li, Z.~Zhang, R.~Wang, Z.~Tu, H.~Mi, and D.~Yu, ``Thoughts are all over the place: On the underthinking of o1-like llms,'' 2025. 


\bibitem{Overthinking_Survey}
Y.~Sui, Y.-N. Chuang, G.~Wang, J.~Zhang, T.~Zhang, J.~Yuan, H.~Liu, A.~Wen, S.~Zhong, H.~Chen, and X.~Hu, ``Stop overthinking: A survey on efficient reasoning for large language models,'' 2025. 

\bibitem{Levin_Tree_Search}
L.~Orseau, L.~Lelis, T.~Lattimore, and T.~Weber, ``Single-agent policy tree search with guarantees,'' \emph{Advances in Neural Information Processing Systems}, vol.~31, 2018.

\bibitem{AR-MCTS}
G.~Dong, C.~Zhang, M.~Deng, Y.~Zhu, Z.~Dou, and J.-R. Wen, ``Progressive multimodal reasoning via active retrieval,'' 2024. 

\bibitem{Mulberry}
H.~Yao, J.~Huang, W.~Wu, J.~Zhang, Y.~Wang, S.~Liu, Y.~Wang, Y.~Song, H.~Feng, L.~Shen, and D.~Tao, ``Mulberry: Empowering mllm with o1-like reasoning and reflection via collective monte carlo tree search,'' 2024. 

\bibitem{AStar}
J.~Wu, M.~Feng, S.~Zhang, R.~Jin, F.~Che, Z.~Wen, and J.~Tao, ``Boosting multimodal reasoning with mcts-automated structured thinking,'' 2025. 

\bibitem{LLamaV-o1}
O.~Thawakar, D.~Dissanayake, K.~More, R.~Thawkar, A.~Heakl, N.~Ahsan, Y.~Li, M.~Zumri, J.~Lahoud, R.~M. Anwer, H.~Cholakkal, I.~Laptev, M.~Shah, F.~S. Khan, and S.~Khan, ``Llamav-o1: Rethinking step-by-step visual reasoning in llms,'' 2025. 

\bibitem{Atom_of_Thoughts}
F.~Teng, Z.~Yu, Q.~Shi, J.~Zhang, C.~Wu, and Y.~Luo, ``Atom of thoughts for markov llm test-time scaling,'' 2025. 

\bibitem{CODEIO}
J.~Li, D.~Guo, D.~Yang, R.~Xu, Y.~Wu, and J.~He, ``Codei/o: Condensing reasoning patterns via code input-output prediction,'' 2025. 

\bibitem{FRE}
J.~N. Farr, J.~J. Jenkins, and D.~G. Paterson, ``Simplification of flesch reading ease formula.'' \emph{Journal of applied psychology}, vol.~35, no.~5, p. 333, 1951.

\bibitem{Canny}
W.~Rong, Z.~Li, W.~Zhang, and L.~Sun, ``An improved canny edge detection algorithm,'' in \emph{2014 IEEE international conference on mechatronics and automation}.\hskip 1em plus 0.5em minus 0.4em\relax IEEE, 2014, pp. 577--582.

\bibitem{VLMEvalkit}
H.~Duan, J.~Yang, Y.~Qiao, X.~Fang, L.~Chen, Y.~Liu, X.~Dong, Y.~Zang, P.~Zhang, J.~Wang \emph{et~al.}, ``Vlmevalkit: An open-source toolkit for evaluating large multi-modality models,'' in \emph{Proceedings of the 32nd ACM International Conference on Multimedia}, 2024, pp. 11\,198--11\,201.

\bibitem{huang2024opera}
Q.~Huang, X.~Dong, P.~Zhang, B.~Wang, C.~He, J.~Wang, D.~Lin, W.~Zhang, and N.~Yu, ``Opera: Alleviating hallucination in multi-modal large language models via over-trust penalty and retrospection-allocation,'' in \emph{Proceedings of the IEEE/CVF Conference on Computer Vision and Pattern Recognition}, 2024, pp. 13\,418--13\,427.


\bibitem{yin2024woodpecker}
S.~Yin, C.~Fu, S.~Zhao, T.~Xu, H.~Wang, D.~Sui, Y.~Shen, K.~Li, X.~Sun, and E.~Chen, ``Woodpecker: Hallucination correction for multimodal large language models,'' \emph{Science China Information Sciences}, vol.~67, no.~12, p. 220105, 2024.


\bibitem{wang2024cogvlm}
W.~Wang, Q.~Lv, W.~Yu, W.~Hong, J.~Qi, Y.~Wang, J.~Ji, Z.~Yang, L.~Zhao, S.~XiXuan, J.~Xu, K.~Chen, B.~Xu, J.~Li, Y.~Dong, M.~Ding, and J.~Tang, ``Cog{VLM}: Visual expert for pretrained language models,'' in \emph{The Thirty-eighth Annual Conference on Neural Information Processing Systems}, 2024. 


\bibitem{yu2024rlaifv}
T.~Yu, H.~Zhang, Y.~Yao, Y.~Dang, D.~Chen, X.~Lu, G.~Cui, T.~He, Z.~Liu, T.-S. Chua, and M.~Sun, ``Rlaif-v: Aligning mllms through open-source ai feedback for super gpt-4v trustworthiness,'' \emph{arXiv preprint arXiv:2405.17220}, 2024.


\bibitem{zhang2023multicot}
Z.~Zhang, A.~Zhang, M.~Li, H.~Zhao, G.~Karypis, and A.~Smola, ``Multimodal chain-of-thought reasoning in language models,'' \emph{arXiv preprint arXiv:2302.00923}, 2023.

\bibitem{mondal2024kamcot}
D.~Mondal, S.~Modi, S.~Panda, R.~Singh, and G.~S. Rao, ``Kam-cot: knowledge augmented multimodal chain-of-thoughts reasoning,'' in \emph{Proceedings of the Thirty-Eighth AAAI Conference on Artificial Intelligence and Thirty-Sixth Conference on Innovative Applications of Artificial Intelligence and Fourteenth Symposium on Educational Advances in Artificial Intelligence}, ser. AAAI'24/IAAI'24/EAAI'24.\hskip 1em plus 0.5em minus 0.4em\relax AAAI Press, 2024. 


\bibitem{yan2025medhalltuneinstructiontuningbenchmarkmitigating}
Q.~Yan, Y.~Yuan, X.~Hu, Y.~Wang, J.~Xu, J.~Li, C.-W. Fu, and P.-A. Heng, ``Medhalltune: An instruction-tuning benchmark for mitigating medical hallucination in vision-language models,'' 2025. 

\bibitem{Xiao_Huang_Gan_He_Li_Yu_Shu_Jiang_Zhu_2025}
W.~Xiao, Z.~Huang, L.~Gan, W.~He, H.~Li, Z.~Yu, F.~Shu, H.~Jiang, and L.~Zhu, ``Detecting and mitigating hallucination in large vision language models via fine-grained ai feedback,'' \emph{Proceedings of the AAAI Conference on Artificial Intelligence}, vol.~39, no.~24, pp. 25\,543--25\,551, Apr. 2025.

\bibitem{SBSC}
K.~Singh, A.~Biswas, S.~Bhowmick, P.~Moturi, and S.~K. Gollapalli, ``Sbsc: Step-by-step coding for improving mathematical olympiad performance,'' 2025. 

\bibitem{DeepSeek}
DeepSeek-AI, D.~Guo, D.~Yang, et.al, ``Deepseek-r1: Incentivizing reasoning capability in llms via reinforcement learning,'' 2025. 

\bibitem{robot_vqa_surgery}
Y.~Du, K.~Chen, Y.~Zhan, C.~H. Low, T.~You, M.~Islam, Z.~Guo, Y.~Jin, G.~Chen, and P.-A. Heng, ``Llm-assisted multi-teacher continual learning for visual question answering in robotic surgery,'' 2024.

\bibitem{automatic_HOI}
Z.~Deng, Y.~Shi, K.~Ji, L.~Xu, S.~Huang, and J.~Wang, ``Human-object interaction via automatically designed vlm-guided motion policy,'' 2025. 

\bibitem{OKVQA}
K.~Marino, M.~Rastegari, A.~Farhadi, and R.~Mottaghi, ``Ok-vqa: A visual question answering benchmark requiring external knowledge,'' 2019.

\bibitem{bai2025hallucinationmultimodallargelanguage}
Z.~Bai, P.~Wang, T.~Xiao, T.~He, Z.~Han, Z.~Zhang, and M.~Z. Shou, ``Hallucination of multimodal large language models: A survey,'' 2025.

\bibitem{2024rlhfv}
T.~Yu, Y.~Yao, H.~Zhang, T.~He, Y.~Han, G.~Cui, J.~Hu, Z.~Liu, H.-T. Zheng, and M.~Sun, ``Rlhf-v: Towards trustworthy mllms via behavior alignment from fine-grained correctional human feedback,'' in \emph{2024 IEEE/CVF Conference on Computer Vision and Pattern Recognition (CVPR)}, 2024, pp. 13\,807--13\,816.

\end{thebibliography}
\end{document}